\documentclass[11pt,letterpaper]{article}
\usepackage{emnlp2017}
\usepackage{times}
\usepackage{latexsym}
\usepackage{listings}
\usepackage{amsmath}
\usepackage{amssymb}
\usepackage{graphicx}

\DeclareFixedFont{\ttb}{T1}{txtt}{bx}{n}{12} 
\DeclareFixedFont{\ttm}{T1}{txtt}{m}{n}{12}  

\usepackage{color}
\definecolor{deepblue}{rgb}{0,0,0.5}
\definecolor{deepred}{rgb}{0.6,0,0}
\definecolor{deepgreen}{rgb}{0,0.5,0}

\definecolor{ipython_cyan}{RGB}{64, 128, 128}
\definecolor{ipython_red}{RGB}{186, 33, 33}
\definecolor{keywords}{RGB}{255,0,90}
\definecolor{comments}{RGB}{0,0,113}
\definecolor{red}{RGB}{160,0,0}
\definecolor{green}{RGB}{0,150,0}
\definecolor{MatlabCellColour}{RGB}{252,251,220}

\newcommand\pythonstyle{\lstset{
language=Python,
basicstyle=\ttm\scriptsize,
otherkeywords={self},             
keywordstyle=\ttb\color{deepblue}\scriptsize\textbf,
emph={MyClass,__init__},          
emphstyle=\ttb\color{deepred},    
stringstyle=\color{deepgreen}\textbf,
commentstyle=\color{ipython_red}\ttfamily\textbf,
frame=tb,                         
showstringspaces=false,     
backgroundcolor = \color{MatlabCellColour},
}}

\lstnewenvironment{python}[1][]
{
\pythonstyle
\lstset{#1}
}
{}

\emnlpfinalcopy

\def\emnlppaperid{***}

\title{\texttt{Function} \texttt{ Assistant}: A Tool for NL Querying of APIs}

\author{Kyle Richardson \and Jonas Kuhn \\
 Institute for Natural Language Processing  \\
  University of Stuttgart \\
  {\tt kyle@ims.uni-stuttgart.de}}

\date{}

\begin{document}

\maketitle

\begin{abstract}
  In this paper, we describe \texttt{\textbf{Function Assistant}}, a lightweight Python-based toolkit for querying and exploring source code repositories using natural language. The toolkit is designed to help end-users of a target API quickly find information about functions through high-level natural language queries and descriptions. For a given text query and background API, the tool finds candidate functions by performing a translation from the text to known representations in the API using the semantic parsing approach of \newcite{Richardson17}. Translations are automatically learned from example text-code pairs in example APIs. The toolkit includes features for building translation pipelines and query engines for arbitrary source code projects. To explore this last feature, we perform new experiments on 27 well-known Python projects hosted on Github. 
  

\end{abstract}

\section{Introduction}

Software developers frequently shift between using different third-party software libraries, or APIs, when developing new applications. Much of the development time is dedicated to understanding the structure of these APIs, figuring out where the target functionality lies, and learning about the peculiarities of how such software is structured or how naming conventions work. When the target API is large, finding the desired functionality can be a formidable and time consuming task. Often developers resort to resources like Google or StackOverflow to find (usually indirect) answers to questions. 


\begin{figure}
\centering
 
\begin{python}
## from nltk.parse.dependencygraph.py

class DependencyGraph(object): 
   """A container ....for  a dependency structure"""
   
   def remove_by_address(self, address):
     """
     Removes the node with the given address.
     """
      # => implementation
        
   def add_arc(self, head_address, mod_address):
     """Adds an arc from the node specified by 
     head_address to the node specified by 
     the mod address....
     """
\end{python}

\caption{Example function documentation in Python NLTK about dependency graphs.}
\end{figure}

We illustrate these issues in Figure 1 using two example functions from the well-known NLTK toolkit. Each function is paired with a short \emph{docstring}, i.e., the quoted description under each function, which provides a user of the software a description of what the function does. While understanding the documentation and code requires technical knowledge of dependency parsing and graphs, even with such knowledge, the function naming conventions are rather arbitrary. The function \texttt{add\_arc} could just as well be called \texttt{create\_arc}. An end-user expecting another naming convention might be left astray when searching for this functionality. Similarly, the available description might deviate from how an end-user would describe such functionality.

Understanding the \texttt{remove\_by\_address} function, in contrast, requires knowing the details of the particular \texttt{DependencyGraph} implementation being used.  Nonetheless, the function corresponds  to the standard operation of \emph{removing a node} from a dependency graph. Here, the technical details about how this removal is specific to a \emph{given address} might obfuscate the overall purpose of the function, making it hard to find or understand.  


At a first approximation, navigating a given API requires knowing \emph{correspondences} between textual descriptions and source code representations. For example, knowing that the English expression \emph{Adds an arc} in Figure 1 translates (somewhat arbitrarily) to \texttt{add\_arc}, or that \emph{given address} translates to \texttt{address}. One must also know how to detect paraphrases of certain target entities or actions, for example that \emph{adding an arc} means the same as \emph{creating an arc} in this context. Other technical correspondences, such as the relation between an \texttt{address} and the target dependency graph implementation, must be learned.


In our previous work (\newcite{Richardson17}, henceforth RK), we look at learning these types of correspondences from example API collections in a variety of programming languages and source natural languages. We treat each given API,  consisting of text and function representation pairs, as a parallel corpus for training a simple semantic parsing model.  In addition to learning translational correspondences, of the type described above, we achieve improvements by adding document-level features that help to learn other technical correspondences. 


In this paper, we focus on using our models as a tool for querying API collections. Given a target API, our model learns an MT-based semantic parser that translates text to code representations in the API. End-users can formulate natural language queries to the background API, which our model will translate into candidate function representations with the goal of finding the desired functionality.  Our tool, called \texttt{\textbf{Function Assistant}} can be used in two ways: as a black-box pipeline for building models directly from arbitrary API collections. As well, it can be customized and integrated with other outside components or models using the tool's flexible internal Python API.  

In this paper, we focus on the first usage of our tool. To explore building models for new API collections, we run our pipeline on 27 open source Python projects from the well-known \emph{Awesome Python} project list.\footnote{\url{github.com/vinta/awesome-python}} As in previous work, we perform synthetic experiments on these datasets, which measure how well our models can generate function representations for unseen API descriptions, which mimic user queries. 






\section{Related Work}


Natural language querying of APIs has long been a goal in software engineering, related to the general problem of software reuse \cite{Krueger}. To date, a number of industrial scale products are available in this area.\footnote{e.g., \url{www.krugle.com,www.searchcode.com}}To our knowledge, most implementations use shallow term matching and/or information-extraction techniques \cite{lv2015codehow}, differing from our methods that use more conventional NLP components and techniques. As we show in this paper and in RK, term matching and related techniques can sometimes serve as a competitive baseline, but are almost always outperformed by our translation approach. 

More recently, there has been increased interest in machine learning on learning code representations from APIs,  especially using resources such as GitHub or StackOverflow.  However, this work tends to look at learning from many API collections \cite{gu2016deep}, making such systems hard to evaluate and to apply to querying specific APIs. Other work looks at learning to generate longer code from source code annotations for natural language programming \cite{allamanis2015bimodal}, often focusing narrowly on a specific programming language (e.g., Java) or set of APIs. To our knowledge, none of these approaches include companion software 
that facilitate building custom pipelines for specific APIs and querying. 


Technically, our approach is related to work on semantic parsing, which looks at generating formal representations from text input for natural language understanding applications, notably question-answering. Many existing methods take direct inspiration from work on MT \cite{wongMAIN} and parsing \cite{ZettlemoyerO}. Please see RK for more discussion and pointers to related work. 


\section{Technical Approach}

In this paper, we focus on learning to generate function representations from textual descriptions inside of source code collections, or APIs. We will refer to these target function representations as \emph{API components}. Each component specifies a function name, a list of arguments, and other optional information such as a namespace. 

Given a set of example text-component pairs from an example API, $D = \{(x_{i},z_{i})\}_{i=1}^{n}$, the goal is to learn how to generate correct, well-formed components $z \in \mathcal{C}$ for each text $x$. When viewed as a semantic parsing problem, we can view each $z$ as analogous to a  target logical form. In this paper, we focus narrowly on Python source code projects, and thus Python functions $z$, however our methods are agnostic to the input natural language and output programming language as shown in RK. 

When used for querying, our model takes a text input and attempts to generate the desired function representation. Technically, our approach follows our previous work and has two components: a simple and lightweight word-based translation model that generates candidate API components, and a discriminative model that reranks the translation model output using additional phrase and document-level features.  All of these models are implemented natively in our tool, and we describe each part in turn. 

\subsection{Translation Model}

Given an input text (or query) sequence $x = w_{1},..,w_{|x|}$, the goal is to generate an output API component $z = u_{i},..,u_{|z|}$, which involves learning a conditional distribution $p(z \mid x)$. We pursue a noisy-channel approach, where 
\begin{equation*}
p( z \mid x) \propto p(x \mid z) p(z)
\end{equation*}

By assuming a uniform prior $p(z)$ on output components, the model therefore involves computing $p(x \mid z)$, which under a word-based translation model can be expressed as: 
\begin{equation*}
p( x \mid z) = \sum_{a} p(x,a \mid z)
\end{equation*}
where  the summation ranges over the set of all many-to-one (word) alignments $a$ from $x \to z$. 

While many different formulations of word-based models exist, we previously found that the simplest lexical translation model, or IBM Model 1 \cite{brown1993mathematics}, outperforms even higher-order alignment models with location parameters. This model computes all alignments exactly using the following equation:
\begin{equation}
 p(x \mid z) \approx \prod_{j=1}^{\mid x\mid} \sum_{i=0}^{\mid z \mid} p_{t}(w_{j} \mid u_{i})
\end{equation}
where $p_{t}$ defines a multinomial distribution over a given component term $u_{j}$ for all words $w_{j}$.

While many parameter estimation strategies exist for training word-based models, we similarly found that the simplest EM procedure of \newcite{brown1993mathematics} works the best. In RK, we describe a linear-time decoding strategy (i.e., for generating components from input) over the number of components $\mathcal{C}$, which we use in this paper. Our tool also implements our types of conventional MT decoding strategies that are better suited for large APIs and more complex semantic languages. 









\subsection{Discriminative Reranking}

Following most semantic parsing approaches \cite{ZettlemoyerO}, we use a discriminative log-linear model to rerank the components generated from the underlying translation model. Such a model defines a conditional distribution:  $ {p (\thinspace z \vert \thinspace x; \theta) \propto e^{\theta \cdotp \phi \left(x,z\right)} }$, for a parameter vector $\theta \in \mathbb{R}^{b}$, and a set of feature functions $\phi(x,z)$. 

Our tool implements several different training and optimization methods. For the purpose of this paper, we train our models using an online, stochastic gradient ascent algorithm under a maximum conditional log-likelihood objective.


\subsubsection{Features} 

For a given text input $x$ and output component $z$, $\phi(x,z)$ defines a set of features between these two items. By default, our pipeline implementation uses three classes of features, identical to the feature set used in RK. The first class includes additional word-level features, such as word/component match, overlap, component syntax information, and so on. The second includes phrase and hierarchical phrase features between text and component candidates, which are extracted standardly from symmetric word-level alignment heuristics. 

The other category of features includes document-level features. This includes information about the underlying API class hierarchy, and relations between words/phrases and abstract classes within this hierarchy. Also, we use additional textual description of parameters in the docstrings to indicate whether word-components candidate pairs overlap in these descriptions. 





\section{Implementation and Usage}

All of the functionality above is implemented in the \texttt{\textbf{Function Assistant}} toolkit. The tool is part of the companion software release for our previous work called \texttt{\textbf{Zubr}}. For efficiency, the core functionality is written in Cython \footnote{\url{http://cython.org/}}, which is a compiled superset of the Python language that facilitates native C/C++ integration. 


The tool is designed to be used in two ways: first,  as a black-box pipeline to build custom translation pipelines and API query engines. The tool can also be integrated with other components using our Cython and Python API. We focus on the first type of functionality.

\subsection{Library Design and Pipelines}

\begin{figure}
\centering
\begin{python}
## pipeline parameters
params = [
  ("-baseline","baseline",False,"bool",
 "Use baseline model [default=False]","GPipeline")
] 

## Zubr pipeline tasks 
tasks = [
  "zubr.doc_extractor.DocExtractor",# extract docs
  "process_data", # custom function.  
  "zubr.SymmetricAlignment",# learn trans. model.
  "zubr.Dataset", # build dataset obj.
  "zubr.FeatureExtractor",  ## build extractor obj.
  "zubr.Optimizer", ## train reranking model 
  "zubr.QueryInterface", # build query interface
  "zubr.web.QueryServer", # launch HTTP server
]

def process_data(config): 
    """Preprocess the extracted data using a custom 
     function or outside library (e.g., nltk)
	
     :param config: The global configuration 
     """ 
     preprocess_function(config,...) 

\end{python}

%

%
%

\caption{An example pipeline script for building a translation model and query server.}
\end{figure}

Our library uses dependency-injection OOP design principles.  All of the core components  are implemented as wholly independent classes, each of which has a number of associated configuration values. These components interact via a class called \texttt{Pipeline}, which glues together various user-specified components and dependencies, and builds a global configuration from these components. Subsequent instantiation and sharing of objects is dictated, or \emph{injected}, by these global configurations settings, which can change dynamically throughout a pipeline run. 

Pipelines are created by writing pipeline scripts, such as the one shown in Figure 2. This file is an ordinary Python file, with two mandatory variables. The first \texttt{params} variable specifies various high-level configuration parameters associated with the pipeline. In this case, there is a setting \texttt{--baseline}, which can be evoked to run a baseline experiment, and will effect the subsequent processing pipeline. 

The second, and most important,  variable is called \texttt{tasks}, and this specifies an ordering of subprocesses that should be executed. The fields in this list are pointers to either core utilities in the underlying \texttt{Zubr} toolkit (each with the prefix \texttt{zubr.}), or user defined functions. This particular pipeline starts by building a dataset from a user specified source code repository, using \texttt{DocExtractor}, then builds a symmetric translation model \texttt{SymmetricAlignment}, a feature extractor \texttt{FeatureExtractor}, a discriminative reranker \texttt{Optimizer}, all via various intermediate steps. It finishes by building a query interface and query server, \texttt{QueryInterface} and \texttt{QueryServer}, which can then be used for querying the input API. 


As noted already, each subprocesses has a number of associated configuration settings, which are joined into a global configuration object by the \texttt{Pipeline} instance. For the translation model, settings include, for example, the type of translation model to use, the number of iterations to use when training models, and so on. All of these settings can be specified on the terminal, or in a separate configuration file. As well, the user is free to define custom functions, such as \texttt{process\_data}, or classes which can be used to modify the default processing pipeline or implement new ML features. 


\subsection{Web Server}

\begin{figure*}
\centering
\framebox(440,219){\includegraphics[scale=.21]{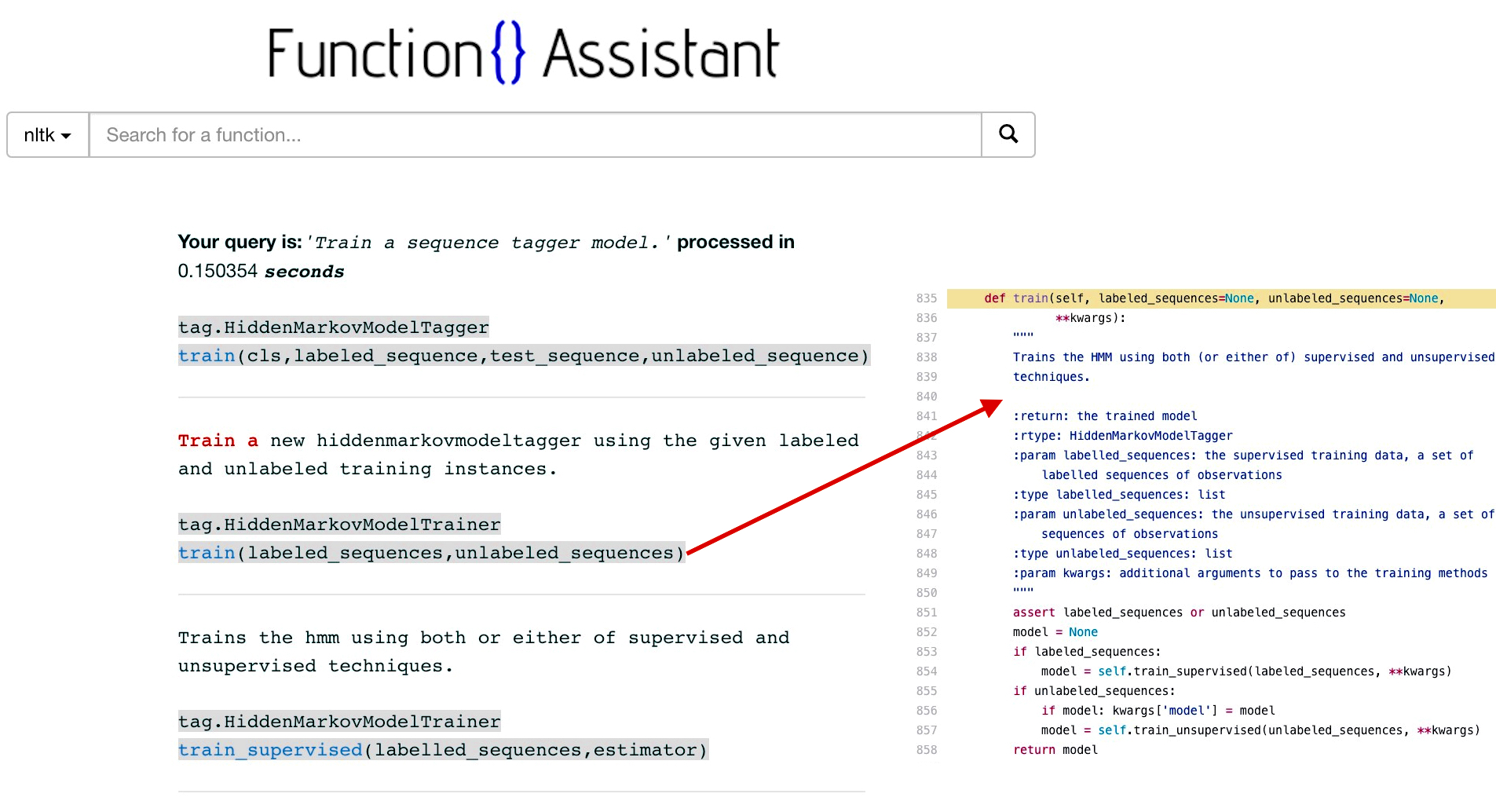}}
\caption{An example screen shot of the \texttt{Function Assistant} web server. }
\end{figure*}

The last step in this pipeline builds an HTTP web server that can be used to query the input API. Internally, the server makes calls to the trained translation model and discriminative reranker, which takes user queries and attempts to translate them into API function representations. These candidate translations are then returned to the user as potential answers to the query. Depending on the outcome, the user can either rephrase his/her question if the target function is not found, or look closer at the implementation by linking to the function's source code. 

An example screen shot of the query server is shown in Figure 3. Here, the background API is the NLTK toolkit, and the query is \emph{Train a sequence tagger model}. While not mentioned explicitly, the model returns training functions for the \texttt{HiddenMarkovModelTagger}. The right side of the image shows the hyperlink path to the original source in Github for the \texttt{train} function.


\begin{table}
   \centering
   \scriptsize 
   \begin{tabular}{| l l l l l |}
   \hline 
   \textbf{Project} & \textbf{\# Pairs} & \textbf{\# Symbols} & \textbf{\# Words} &\textbf{Vocab.} \\
   \hline \hline
   		\texttt{scapy} & 757 & 1,029 & 7,839 & 1,576 \\ 
		\texttt{zipline} & 753 & 1,122 & 8,184 & 1,517 \\ 
		\texttt{biopython} & 2,496 & 2,224 & 20,532 & 2,586 \\ 
		\texttt{renpy} & 912 & 889 & 10,183 & 1,540 \\ 
		\texttt{pyglet} & 1,400 & 1,354 & 12,218 & 2,181 \\ 
		\texttt{kivy} & 820 & 861 & 7,621 & 1,456 \\ 
		\texttt{pip} & 1,292 & 1,359 & 13,011 & 2,201 \\ 
		\texttt{twisted} & 5,137 & 3,129 & 49,457 & 4,830 \\ 
		\texttt{vispy} & 1,094 & 1,026 & 9,744 & 1,740 \\ 
		\texttt{orange} & 1,392 & 1,125 & 11,596 & 1,761 \\ 
		\texttt{tensorflow} & 5,724 & 4,321 & 45,006 & 4,672 \\ 
		\texttt{pandas} & 1,969 & 1,517 & 17,816 & 2,371 \\ 
		\texttt{sqlalchemy} & 1,737 & 1,374 & 15,606 & 2,039 \\ 
		\texttt{pyspark} & 1,851 & 1,276 & 18,775 & 2,200 \\ 
		\texttt{nupic} & 1,663 & 1,533 & 16,750 & 2,135 \\ 
		\texttt{astropy} & 2,325 & 2,054 & 24,567 & 3,007 \\ 
		\texttt{sympy} & 5,523 & 3,201 & 52,236 & 4,777 \\ 
		\texttt{ipython} & 1,034 & 1,115 & 9,114 & 1,771 \\ 
		\texttt{orator} & 817 & 499 & 6,511 & 670 \\ 
		\texttt{obspy} & 1,577 & 1,861 & 14,847 & 2,169 \\ 
		\texttt{rdkit} & 1,006 & 1,380 & 9,758 & 1,739 \\ 
		\texttt{django} & 2,790 & 2,026 & 31,531 & 3,484 \\ 
		\texttt{ansible} & 2,124 & 1,884 & 20,677 & 2,593 \\ 
		\texttt{statsmodels} & 2,357 & 2,352 & 21,716 & 2,733 \\ 
		\texttt{theano} & 1,223 & 1,364 & 12,018 & 2,152 \\ 
		\texttt{nltk} & 2,383 & 2,324 & 25,823 & 3,151 \\ 
		\texttt{sklearn} & 1,532 & 1,519 & 13,897 & 2,115 \\ 

   \hline
   \end{tabular}

\caption{New English Github datasets.}
\end{table}

\section{Experiments}

Our current \texttt{DocExtractor} implementation supports building parallel datasets from raw Python source code collections. Internally, the tool reads source code using the abstract syntax tree utility, \texttt{ast}, in the Python standard library, and extracts sets of function and description pairs.  In addition, the tool also extracts class descriptions, parameter and return value descriptions, and information about the API's internal class hierarchy. This last type of information is then used to define document-level features. 

To experiment with this feature, we built pipelines and ran experiments for 27 popular Python projects. The goal of these experiments is to test the robustness of our extractor, and see how well our models answer unseen queries for these resources using our previous experimental setup. 




\subsection{Datasets} 

The example projects are shown in Table 1. Each dataset is quantified in terms of \textbf{\# Pairs}, or the number of parallel function-component representations, the \textbf{\# Symbols} in the component output language, the \textbf{\#} (NL) \textbf{Words} and \textbf{Vocab} size. 

\begin{table*}
\centering
\scriptsize
\setlength{\tabcolsep}{2pt}

    \begin{tabular}{l c c c c c c c c c}
    \hline\hline
    \textbf{Method} &   \texttt{scapy}  &  \texttt{zipline}  &  \texttt{biopython}  &  \texttt{renpy}  &  \texttt{pyglet}  &  \texttt{kivy}  &  \texttt{pip}  &  \texttt{twisted}  &  \texttt{vispy}   \\ \hline
    \textbf{BoW} &  00.0 51.3 17.4 & 01.7 38.3 12.9 & 05.8 54.8 20.4 & 06.6 41.1 16.6 & 05.7 52.3 19.2 & 07.3 53.6 22.0 & 06.2 40.9 17.1 & 06.6 38.8 16.9 & 07.3 48.7 18.6  \\
    \textbf{Term Match} &  \textbf{21.2}  43.3 28.7 & 28.5 50.8 36.2 & 23.5 48.1 31.7 & 25.7 59.5 38.7 & 20.4 50.9 31.2 & 30.0 62.6 41.3 & 19.1 50.2 30.7 & 17.6 44.1 26.2 & 29.2 64.0 41.1  \\
    \textbf{Translation} & 20.3 61.9 34.7 & 27.6 62.5 40.7 & 29.6 75.6 45.8 & 30.8 61.7 42.0 & 26.1 69.5 41.3 & 33.3 67.4 45.3 & 18.6 56.4 32.3 & 27.7 61.4 39.4 & 28.6 70.1 42.3  \\
    \textbf{Reranker} &  \textbf{21.2}  \textbf{67.2}   \textbf{37.2}  &  \textbf{30.3}   \textbf{70.5}   \textbf{45.3}  &  \textbf{32.3}   \textbf{79.1}   \textbf{48.6}  &  \textbf{38.9}   \textbf{73.5}   \textbf{48.9}  &  \textbf{29.0}   \textbf{77.1}   \textbf{45.5}  &  \textbf{35.7}   \textbf{75.6}   \textbf{49.1}  &  \textbf{25.9}   \textbf{65.8}   \textbf{39.9}  &  \textbf{28.8}   \textbf{65.8}   \textbf{42.2}  &  \textbf{33.5}   \textbf{80.4}   \textbf{50.3}   \\
    \end{tabular}

    \begin{tabular}{l c c c c c c c c c}
    \hline\hline
    \textbf{Method} &   \texttt{orange}  &  \texttt{tensorflow}  &  \texttt{pandas}  &  \texttt{sqlalchemy}  &  \texttt{pyspark}  &  \texttt{nupic}  &  \texttt{astropy}  &  \texttt{sympy}  &  \texttt{ipython}   \\ \hline
    \textbf{BoW} &  13.4 60.5 29.1 & 09.4 47.4 21.2 & 03.7 40.6 15.6 & 07.3 45.0 18.4 & 07.5 50.9 20.8 & 06.4 55.0 22.8 & 07.7 52.0 21.1 & 06.4 44.4 18.5 & 01.9 41.2 13.9  \\
    \textbf{Term Match} &  37.9 69.7 49.3 & 25.2 48.7 33.5 & 19.3 43.7 27.9 & 17.3 48.4 26.6 & 20.5 46.9 29.1 & 23.6 51.0 33.1 & 26.1 49.1 34.3 & 20.2 44.9 28.8 & 23.8 56.7 33.8   \\
    \textbf{Translation} &  40.3 78.3 54.0 & 35.3 71.5 48.0 & 29.1 62.7 41.0 & 28.8 70.3 43.0 & 37.1 78.7 52.1 &  \textbf{30.9}  69.8 44.6 & 30.7 66.6 43.4 &  \textbf{32.8}  70.2 45.5 & 24.5 59.3 36.5  \\
    \textbf{Reranker} &   \textbf{45.1}   \textbf{84.1}   \textbf{59.9}  &  \textbf{38.4}   \textbf{77.7}   \textbf{51.8}  &  \textbf{31.1}   \textbf{66.1}   \textbf{43.1}  &  \textbf{35.0}   \textbf{76.1}   \textbf{49.7}  &  \textbf{41.5}   \textbf{81.5}   \textbf{55.3}  & 29.3  \textbf{76.7}   \textbf{45.6}  &  \textbf{33.9}   \textbf{74.4}   \textbf{47.4}  & 32.1  \textbf{75.0}   \textbf{46.6}  &  \textbf{29.6}   \textbf{66.4}   \textbf{42.3}   \\
    \end{tabular}

    \begin{tabular}{l c c c c c c c c c}
    \hline\hline
    \textbf{Method} &   \texttt{orator}  &  \texttt{obspy}  &  \texttt{rdkit}  &  \texttt{django}  &  \texttt{ansible}  &  \texttt{statsmodels}  &  \texttt{theano}  &  \texttt{nltk}  &  \texttt{sklearn}   \\ \hline
    \textbf{BoW} &  10.6 66.3 28.6 & 06.7 49.5 20.2 & 05.3 40.6 17.1 & 04.5 40.9 16.2 & 17.9 55.3 30.5 & 05.6 46.1 18.6 & 03.2 43.7 16.2 & 05.0 44.2 16.3 & 05.2 45.8 17.7  \\
    \textbf{Term Match} &  31.9 64.7 43.7 & 19.9 46.6 30.0 & 13.3 46.6 23.9 & 19.3 48.0 29.1 & 24.8 54.0 35.8 & 16.7 39.9 25.1 & 16.3 37.1 24.0 & 19.8 45.6 28.4 & 24.4 50.6 32.5  \\
    \textbf{Translation} &   \textbf{32.7}  79.5 47.5 & 33.8 75.8 48.3 &  \textbf{25.3}  60.6 37.2 & 22.9 57.8 34.6 & 35.5 71.6 47.5 & 25.4 64.8 37.8 & 26.2 58.4 37.8 & 28.2 68.0 41.5 & 27.9 67.6 41.3  \\
    \textbf{Reranker} &  \textbf{32.7}  \textbf{82.7}   \textbf{49.7}  &  \textbf{37.7}   \textbf{80.0}   \textbf{52.3}  & \textbf{25.3}  \textbf{63.3}   \textbf{39.6}  &  \textbf{25.8}   \textbf{64.5}   \textbf{39.4}  &  \textbf{40.5}   \textbf{77.0}   \textbf{53.1}  &  \textbf{28.8}   \textbf{69.1}   \textbf{41.7}  &  \textbf{27.3}   \textbf{66.1}   \textbf{39.9}  &  \textbf{31.6}   \textbf{72.5}   \textbf{45.7}  &  \textbf{29.2}   \textbf{75.5}   \textbf{44.5}   \\
    \end{tabular}

\begin{tabular}{| c | c | c |}
\textbf{Accuracy @1} & \textbf{Accuracy @10} & \textbf{Mean Reciprocal Rank (MRR)} \
\end{tabular}

\caption{Test results on our new Github datasets.}
\end{table*}

\subsection{Experimental Setup}

Each dataset is randomly split into train, test, and dev. sets using a 70\%-30\% (or 15\%/15\%) split. We can think of the  held-out sets as mimicking queries that users might ask the model. Standardly, all models are trained on the training sets, and hyper-parameters are tuned to the dev. sets. 

For a unseen text input during testing, the model generates a candidate list of component outputs. An output is considered correct if it matches \emph{exactly} the gold function representation. As before, we measure the \textbf{Accuracy @1}, accuracy within top ten (\textbf{Accuracy @10}), and the \textbf{MRR}. 

As in our previous work, three additional baselines are used. The first is a simple bag-of-words (\textbf{BoW}) model, which uses word-component pairs as features. The second is a \textbf{Term Match} baseline, which ranks candidates according to the number of matches between input words and component words. We also compare the results of the \textbf{Translation} (model) without the reranker. 


\section{Results and Discussion}

Test results are shown in Table 2, and largely conform to our previous findings. The \textbf{BoW} and \textbf{Term Match} baselines are outperformed by all other models, which again shows that API querying is more complicated than simple word-component matching. The \textbf{Reranker} model leads to improvements on all datasets as compared with only using the \textbf{Translation} model, indicating that document-level and phrase features can help. 

We note that these experiments are synthetic, in the sense that it's unclear whether the held-out examples bear any resemblance to actual user queries. Assuming, however, that each held-out set is a representative sample of the queries that real users would ask, we can then interpret the results as indicating how well our models answer queries. Whether or not these held-out examples reflect \emph{real} queries, we believe that they still provide a good benchmark for model construction. All code and data will be released to facilitate further experimentation and application building. Future work will look at eliciting more naturalistic queries (e.g., through StackOverflow), and doing usage studies via a permanent web demo\footnote{{\scriptsize see demo here: \url{http://zubr.ims.uni-stuttgart.de/}}}.  


\section{Conclusion}

We introduce \textbf{Function Assistant}, a lightweight tool for querying API collections using unconstrained natural language. Users can supply our tool with target source code projects and build custom translation or processing pipelines and query servers from scratch. In addition to the tool, we also created new resources for studying API querying, in the form of datasets built from 27 popular Github projects. While our approach uses simple components, we hope will that our tool and resources will serve as a benchmark for future work in this area, and ultimately help to solve everyday software search and reusability issues. 


\bibliography{emnlp2017}

\newcommand{\noop}[1]{}
\begin{thebibliography}{8}
\expandafter\ifx\csname natexlab\endcsname\relax\def\natexlab#1{#1}\fi

\bibitem[{Allamanis et~al.(2015)Allamanis, Tarlow, Gordon, and
  Wei}]{allamanis2015bimodal}
Miltiadis Allamanis, Daniel Tarlow, Andrew~D Gordon, and Yi~Wei. 2015.
\newblock Bimodal modelling of source code and {NL}.
\newblock In \emph{Proceedings of ICML}.

\bibitem[{Brown et~al.(1993)Brown, Pietra, Pietra, and
  Mercer}]{brown1993mathematics}
Peter~F Brown, Vincent J~Della Pietra, Stephen A~Della Pietra, and Robert~L
  Mercer. 1993.
\newblock The mathematics of {SMT}.
\newblock \emph{Computational linguistics}, 19(2).

\bibitem[{Gu et~al.(2016)Gu, Zhang, Zhang, and Kim}]{gu2016deep}
Xiaodong Gu, Hongyu Zhang, Dongmei Zhang, and Sunghun Kim. 2016.
\newblock Deep {API} {L}earning.
\newblock \emph{arXiv preprint arXiv:1605.08535}.

\bibitem[{Krueger(1992)}]{Krueger}
Charles~W. Krueger. 1992.
\newblock Software reuse.
\newblock \emph{ACM Computing Surveys (CSUR)}, 24(2).

\bibitem[{Lv et~al.(2015)Lv, Zhang, Lou, Wang, Zhang, and Zhao}]{lv2015codehow}
Fei Lv, Hongyu Zhang, Jian-guang Lou, Shaowei Wang, Dongmei Zhang, and Jianjun
  Zhao. 2015.
\newblock Codehow: Effective code search based on api understanding and
  extended boolean model (e).
\newblock In \emph{Proceedings of ASE}.

\bibitem[{Richardson and Kuhn(2017)}]{Richardson17}
Kyle Richardson and Jonas Kuhn. 2017.
\newblock Learning {S}emantic {C}orrespondences in {T}echnical {D}ocumentation.
\newblock In \emph{Proceedings of ACL}.

\bibitem[{Wong and Mooney(2006)}]{wongMAIN}
Yuk~Wah Wong and Raymond~J. Mooney. 2006.
\newblock Learning for {S}emantic {P}arsing with {S}tatistical {M}achine
  {T}ranslation.
\newblock In \emph{Proceedings of HLT-NAACL}.

\bibitem[{Zettlemoyer and Collins(2009)}]{ZettlemoyerO}
Luke~S. Zettlemoyer and Michael Collins. 2009.
\newblock Learning context-dependent mappings from sentences to logical form.
\newblock In \emph{Proceedings of ACL}.

\end{thebibliography}


\begin{thebibliography}{4}
\expandafter\ifx\csname natexlab\endcsname\relax\def\natexlab#1{#1}\fi

\bibitem[{Aho and Ullman(1972)}]{Aho:72}
Alfred~V. Aho and Jeffrey~D. Ullman. 1972.
\newblock \emph{The Theory of Parsing, Translation and Compiling}, volume~1.
\newblock Prentice-Hall, Englewood Cliffs, NJ.

\bibitem[{{American Psychological Association}(1983)}]{APA:83}
{American Psychological Association}. 1983.
\newblock \emph{Publications Manual}.
\newblock American Psychological Association, Washington, DC.

\bibitem[{Chandra et~al.(1981)Chandra, Kozen, and Stockmeyer}]{Chandra:81}
Ashok~K. Chandra, Dexter~C. Kozen, and Larry~J. Stockmeyer. 1981.
\newblock \href {https://doi.org/10.1145/322234.322243} {Alternation}.
\newblock \emph{Journal of the Association for Computing Machinery},
  28(1):114--133.

\bibitem[{Gusfield(1997)}]{Gusfield:97}
Dan Gusfield. 1997.
\newblock \emph{Algorithms on Strings, Trees and Sequences}.
\newblock Cambridge University Press, Cambridge, UK.

\end{thebibliography}
\bibliographystyle{emnlp_natbib}

\end{document}